# 半监督学习理论及其研究进展概述

屠恩美，杨　杰

（上海交通大学 电子信息与电气工程学院，上海 200240）

**摘　要**：半监督学习介于传统监督学习和无监督学习之间，是一种新型机器学习方法，其思想是在标记样本数量很少的情况下，通过在模型训练中引入无标记样本来避免传统监督学习在训练样本不足（学习不充分）时出现性能（或模型）退化的问题。半监督学习已在许多领域被成功应用。回顾了半监督学习的发展历程和主要理论，并介绍了半监督学习研究的最新进展，最后结合应用实例分析了半监督学习在解决实际问题中的重要作用。

# A Review of Semi-Supervised Learning Theories and Recent Advances

TU Enmei, Yang Jie

(School of Electronics, Information & Electrical Engineering, Shanghai Jiao Tong University, Shanghai 200240, China)

**Abstract:** Semi-supervised learning, which has emerged from the beginning of this century, is a new type of learning method between traditional supervised learning and unsupervised learning. The main idea of semi-supervised learning is to introduce unlabeled samples into model training process to avoid performance (or model) degeneration due to insufficiency of labeled samples. Semi-supervised learning has been applied sucessfully in many fields. This paper reviews the development process and main theories of semi-supervised learning, as well as its recent advances and importance in solving real-world problems demonstrated by typical application examples.

传统的监督学习（如支持向量机，神经网络等）通常需要大量的良好标记样本对模型进行仔细地训练, 以便获得较好的模型泛化能力。同时，由于维度灾难的问题，在处理高维数据（如视频、语音、图像分类、文档）时，训练一个好的监督模型所需要的标记样本数量会进一步呈现指数暴增趋势。这使得传统监督学习很难应用于一些缺乏训练样本的任务中。例如，医学诊断中某些疾病患者数量较少，且部分患者可能有隐私考虑不愿信息被采集使用；再如，具有破坏性的实验中（汽车碰撞实验、火箭发射、新产品耐用性测试等）也很难收集到大量的训练样本，因为成本很高。因此，在模型的训练中，如何降低学习模型对标记样本数据量的需求，同时提高模型学习的性能，成为一个重要的研究问题。

半监督学习（Semi-supervised Learning）是近十多年来发展起来的一类新型机器学习方法，其思想是在标记样本数量很少的情况下，通过在模型训练中引入无标记样本来避免传统监督学习在训练样本不足（学习不充分）时出现性能（或模型）退化的问题。半监督学习的研究具有重要的实用价值，因为在许多实际应用中，一方面标记样本的获取成本往往较高（如需要花费人力和时间进行标记，需要用到特殊仪器或设备进行实验和测量等）；另一方面无标记样本的获取相对容易，只需要简单重复地采集即可大量收集。因此，在实际应用中减少标记样本的使用能够大幅缩减人力、时间和资源的开销，从而降低生产成本。同时在标记样

本数量减少数十或数百倍（甚至更多）的情况下，算法能够取得与传统大量标记样本训练的监督学习算法相近甚至更好的效果，提升了生产效率。半监督学习的研究也具有重要的理论价值，它是介于传统监督学习（只利用标记样本学习）和无监督学习（只利用无标记样本学习）之间的一种新型机器学习方法，是对传统机器学习理论的拓展和补充。此外，已有的研究表明人类的学习过程中也存在大量的半监督学习行为，因此半监督学习也对认知科学的发展起到了一定的促进作用。

本文首先回顾了半监督学习的发展历程，然后介绍了半监督学习的主要理论和算法，之后详细介绍了笔者实验室在半监督学习研究方面的最新进展和成果，结合实例分析了半监督学习在解决实际问题中的重要作用，最后对半监督学习研究进行了总结和展望。与以往的综述[1, 2]相比，本文总结了半监督学习的发展阶段性特点，并对每个阶段的代表性算法进行了较为全面的综述，另外还包含了其他综述中未曾涉及的最新的半监督深度学习算法。

# 1 半监督学习的发展综述

20 世纪 90 年代就有学者开始尝试在训练分类器时利用无标记样本来提高分类器性能，但 2000 年后半监督学习才逐步形成相对独立的理论和算法体系，成为有别于传统监督学习和无监督学习的一类新型机器学习方法。总体来看，半监督学习大概具有以下三个发展阶段特征：第一个阶段是在上世纪 90 年代期间，人们初步探索无标记样本在一些传统机器学习算法中的作用，是半监督学习的酝酿阶段；第二阶段是本世纪初的十多年间，形成了独立的半监督学习算法和理论体系，是经典半监督学习发展趋于成熟阶段；第三个阶段是最近几年，由于深度学习的巨大成功和普及应用的需求，半监督学习与深度神经网络结合形成的半监督深度学习成为研究热点，把半监督学习研究推向新的高度。下面我们对每个阶段的代表性算法进行介绍。

## 1.1 半监督学习的早期阶段

早期的半监督学习是初步探索无标记样本在传统监督学习模型中的价值[3]，学习算法多数是对传统的机器学习算法进行改进，通过在监督学习中加入无标记样本来实现。这类算法有最大似然分类器[4, 5]，贝叶斯分类器 (Bayes Classifier)[6, 7]，多层感知器[8]，支持向量机[9, 10]等。早期半监督学习算法中有较大影响力的是半监督支持向量机(Semi-supervised Support Vector Machine, $S^3$VM)和协同训练(Co-training)。

$S^3$VM 源于 Vapnik20 世纪 90 年代末的直推式支持向量机(Transductive SVM)研究[11]：在给定较少标记的训练样本情况下，支持向量机的决策边界(Decision Boundary)不应该穿过样本密度高的区域。因此，$S^3$VM 的目标函数是在传统的支持向量机目标函数基础上，增加了一个包含无标记样本的约束项来惩罚分类超平面穿过样本密集区域的情况。修改后的目标函数是一个非凸的组合优化问题，直接求解难度较大且计算量会随着数据集的增大而指数暴增，因此初期的求解算法基本很难在实际应用中使用。Joachims 提出了基于标记切换的组合优化算法[12]，第一次使 $S^3$VM 在具有实用意义的数据集上取得了很好的效果。之后更多的学者开始关注 $S^3$VM，相继出现了不同的求解方法，其中较有影响的算法包括半正定规划(Semi-definite Programming)[13-16]算法，凹凸过程(Concave Convex Procedure)[17-19]，延续法(Continuation)[20]， 梯度下降[21]，确定性退火(Deterministic Annealing) [22]等。另外，为了避免算法把所有样本放在决策边界的同一边这种无意义的求解，文献[12, 20, 21]中还探讨了施加平衡约束(Balancing Constraint)情况下的 $S^3$VM 问题求解。

协同训练[23]假设数据本身具有多个相互独立的视角（Multi-view），且每个视角都可以独立对数据进行分类。针对数据的每个视角，协同训练首先用标记样本训练一个该视角对应的分类器，然后不同视角的分类器对无标记样本进行分类，每个视角的分类器都把自己认为可靠度较高的无标记样本连同其对应的分类标记加入到其他视角分类器的训练数据集中，最后所有分类器利用各自更新后的训练集进行二次训练，如此重复直到完成分类。文献[24]中进一步分析了协同训练能够取得好的效果的原因，同时讨论了协同训练假设不成立时的情况。文献[25]中提出了弱化假设的协同训练，并进行了理论分析。协同学习在自然语言处理中有着重要的的应用，包括语意解析[26]，语意标注[27]，词语的同指解析[28]，歧义去除[29]，跨语言情感分类[30]等。

## 1.2 半监督学习的成熟阶段

由于 $S^3VM$ 是非凸离散组合优化，求解难度大且很难获得全局最优解，同时协同训练假设条件苛刻等，人们开始尝试其他方法进行半监督学习。2000 年后的十多年间大量的半监督学习算法开始涌现，这时期的标志是“半监督学习”的概念被明确提出并形成了崭新的算法体系，使半监督学习逐渐形成相对独立的、区别于传统监督学习和无监督学习的一类学习方法。这个时期的半监督学习主要包括混合模型(Mixture Model)[31-34]、伪标记或自训练(Pseudo Label, Self-training)[29, 35-37]、图论半监督学习[38-40]、流形半监督学习[41-43]等。与其他类型的半监督学习算法相比，基于图论的半监督学习算法有许多优势，如算法多数为凸优化，可求得全局最优解，有些甚至具有闭合数学表达式；算法基于矩阵运算操作，效率较高且便于理解和实现；图论是数学的一个分支，具有很好的理论基础等，因此受到了广泛的关注。

图论半监督学习需要首先构建一个图 (Graph), 图的节点集就是所有样本集（包括标记样本和无标记样本），图的边是样本两两间的相似性（通常使用高斯核函数作为相似性度量），然后把分类问题看作是类别信息在图上由标记节点向无标记节点的扩散或传播过程。代表性的算法包括：文献[38]利用图论中的最小切割(Min-cut)算法来对无标记样本进行分类，文献[44]中把学习过程看作是样本的类别标记在图上的不断扩散和传播，文献[39]中通过求解高斯场谐函数(Gaussian Field Harmonic Function, GFHF)来实现对图上的无标记样本进行分类，稍后文献[40]中提出基于网络的动态传播思想的局部和全局一致性(Local and Global Consistency, LGC)半监督学习算法。GFHF 和 LGC 两个算法都具有很好的理论框架，以及具有闭合解析式的全局最优解，同时算法易于实现，在实际应用中效果突出，因此引起了学者们的广泛关注，成为图论半监督学习的经典算法代表。二者同属于直推式学习(Transductive Learning)方式，即算法只能在参与学习过程的数据集上学习出一个模型，因此对于后续的新样本无法直接给出分类。但二者相比，LGC 算法对样本的标记具有更好的容错性（因为采用的是软约束，而不像 GFHF 中强制的硬性约束），而且实际应用时效果也往往更好。文献[45]中把这两个算法统一在一个理论模型中，并在此基础上提出了基于 Nystrom 核矩阵近似的快速算法，以解决图论半监督学习在大数据集上计算复杂度高、内存消耗大的问题。文献[46]在 LGC 算法的基础上采用了局部线性重构的方式构建图，提出线性邻域标记传播(Linear Neighborhood Label Propagation)算法，受到较多关注。因为图论半监督学习需要构建一个包含所有样本的图，而且求解算法时间和空间复杂度通常是样本的三次方，这带来的一个难题是如何在大数据集上进行学习，文献[47, 48]中对超大型的数据集上如何进行半监督学习做了研究。需要说明的是，图论半监督学习中还有许多其他算法和相关研究，在此无法一一列举。例如，文献[49]中分析了图的构建方式对算法性能的影响并提出了一种 *b*-Matching 图提升算法精度，文献[50]中研究了基于有向图的半监督学习，文献[51]中研究了多图组合下的半监督学习，文献[52]中研究了半监督的降维算法，文献[53]中探讨了参数学习问题等等。

半监督学习研究中，与图论密切相关的另一个理论是微分流形理论。在一定条件下，图论中的拉普拉斯矩阵可以看成是流形上的拉普拉斯-贝尔特拉米算子(Laplace-Beltrami Operator)的离散化[54-56]。因此，以图论作为工具，基于微分流形理论的半监督学习算法也受到广泛关注。这其中的代表算法是 Belkin 等提出的流形正则化算法(Manifold Regularization)[41, 42, 57]，其目标是在高维的数据空间中，通过惩罚分类函数在样本所分布的低维流形上的复杂度来实现无标记样本的利用。其他基于微分流形理论的半监督学习算法包括黎曼流形(Riemannian Manifold)半监督学习[43]，海森正则化(Hessian Regularization)半监督学习[58]，局部坐标编码(Local Coordinate Coding)半监督学习[59]，多流形半监督学习[60]等。

## 1.3 半监督深度学习的发展

随着深度学习在图像识别[61]、自然语言处理[62]和语音识别[63]等方面取得突破，半监督深度学习算法研究就成了自然的需求，因为深度学习普及应用的障碍之一就是对海量标记样本的需求在很多应用中难以被满足。较早进行半监督深度学习研究的是 Weston 等[64]，他们尝试把图论半监督学习中的拉普拉斯正则项引入到神经网络的目标函数中，对多层神经网络进行半监督训练。总结起来，已有的半监督深度学习算法可归为三类：无监督特征学习类，正则化约束类和生成式对抗网络(Generative Adversarial Nets, GAN)类。

无监督特征学习类算法通常利用所有样本（包含标记样本和无标记样本）学习出样本的

隐特征或隐含变量表示(Latent Feature or Hidden Variable)，在此基础上利用有监督分类器对无标记样本所对应的隐特征进行分类，从而间接地对无标记样本进行分类。文献[65]中采用叠加的生成模型来学习标记样本和无标记样本的隐变量并使用 SVM 对学习的隐变量进行分类。文献[66]中首先采用局部区域卷积(Local Region Convolution)在无标记的文本中学习出双视嵌入(Two-View (TV) Embedding)特征，然后采用卷积神经网络进行分类。随后文献[67]中又对该算法进行了拓展，采用 LSTM (Long Short-Term Memory)进行区域大小可变的文本特征学习。文献[68]中把自编码器(Auto-Encoder)的编码层和解码层之间加入短路连接，然后使用分类器对自编码器学习的特征进行分类。文献[69]中把自编码器按顺序拼接在一起，通过最小化这些自编码器的重构误差可以学习出序列数据的隐特征。其他算法还包括弱监督下的激活图学习[70]，图卷积网络学习隐特征[71]。

正则化约束类算法通常是在有监督神经网络的输出层或者隐含层的目标函数中加入体现样本分布特性的正则化项，用以在训练中引入无标记样本。文献[64]中把图的拉普拉斯正则化项分别加入到网络输出层的目标函数和中间隐含层的目标函数中，用来做半监督的分类和特征学习。文献[72]中定义一组标准的随机变换操作，然后定义网络目标函数包括两个部分：监督学习损失函数为标记样本多次随机变换后的预测差异，正则化项为无标记样本多次通过网络预测的结果差异，最后通过反向梯度传播来最小化目标函数进行半监督深度学习。文献[73, 74]中分别使用最大似然分类器和多层感知器作为监督学习的损失函数，并借用自然语言处理中用于词语特征学习的 Skipgram 模型作为正则化项。文献[75]中采用分类指示向量互斥原则对网络进行正则化，即所有样本通过网络后输出的类别向量中只有一个为非 0，这就迫使网络在训练时对无标记样本的分布进行学习并给出确定的类别。其他的正则化半监督学习还包括信息熵正则项(Entropy Regularization)[76, 77]，自编码器正则化[78]，邻域距离正则化[79]。需要说明的是，早期训练深度神经网络时常用的 Pre-training, Fine-tuning 训练方法，如文献[80, 81]，也可看作是一种特殊的正则化[82]。

生成式对抗网络 GAN[83]中，通过让生成器(Generator)和判决器(Discriminator)相互竞争达到平衡状态来无监督地训练网络。由于 GAN 在生成模拟真实样本上的成功表现（如文献[84]中），一个很自然的想法就是在标记样本较少的情况下，能否利用 GAN 所学到的样本内容分布和强大的竞争学习能力来提高网络分类性能。文献[85]中迫使判决器对于真实样本输出单热向量(One-hot Vector)，而对于生成样本输出均匀向量（即类别不确定）。文献[86]中提出了一种输出分布匹配(Output Distribution Matching, ODM)方法用作半监督学习中的正则化项，并用 GAN 对网络进行训练使得生成的虚拟样本类别分布与真实样本类别分布相匹配。文献[87, 88]中通过对判决器进行修改使其输出 $K+1$ 类，其中前 $K$ 类为真实数据的分类，后一类为生成样本分类。这样在 GAN 的训练过程中，分类器需要在判断真假样本的基础上，进一步给出真样本的类别，这就可以借助 GAN 在训练中学到的样本内容分布加上少量标记样本来完成半监督学习。另外，这里强制对样本进行分类与文献[85]中强制输出单热向量目的上是一致的，都是要求判决器尽可能地确定每个样本的类别信息。文献[89]中证明了在半监督学习情况下，一个差的生成器能够更有利于判决器进行半监督地学习，并以此为基础对 GAN 做了修改，通过最小化生成器真假样本分布的 KL 散度和最大化判决器的条件熵来交替训练网络，效果改进明显。

## 2 半监督学习的研究进展

随着半监督学习在许多应用中取得成功，半监督学习的研究也吸引了大批学者，发展迅速。这里简要介绍下笔者所在实验室在图论半监督学习方面的研究进展和成果。

### 2.1 基于泰勒展式的快速半监督学习算法

图论半监督学习（如 LGC 算法）的计算复杂度普遍为样本数 $n$ 的立方级 $O(n^3)$ ，这给大数据集上的应用带来了很大挑战。基于泰勒展式的快速半监督学习算法[90]就是针对 LGC 中的高斯核矩阵进行解析近似，然后利用快速矩阵求逆算法降低计算量，把计算复杂度由立方转变为近似线性。算法的主要思想是首先把高斯核函数分解为三项：

$$\exp(-\frac{\|\boldsymbol{x}_i-\boldsymbol{x}_j\|^2}{2\sigma^2})=\exp(-\frac{\|\boldsymbol{x}_i\|^2}{2\sigma^2})\exp(-\frac{\|\boldsymbol{x}_j\|^2}{2\sigma^2})\exp(\frac{\boldsymbol{x}_i^{\mathrm{T}}\boldsymbol{x}_j}{\sigma^2}) \tag{1}$$

式中：$\boldsymbol{x}_i$ 表示第 $i$ 个样板的特征向量；$\sigma$ 为高斯核函数的方差。然后对最后的点积项利用多变量函数 $f(\boldsymbol{x})$ 泰勒展式进行二阶近似：

$$\begin{aligned} f(\boldsymbol{x}) = {} & f(\boldsymbol{x}_0)+\nabla f(\boldsymbol{x}_0)^{\mathrm{T}}(\boldsymbol{x}-\boldsymbol{x}_0)+ \\ & \frac{1}{2}(\boldsymbol{x}-\boldsymbol{x}_0)^{\mathrm{T}}\boldsymbol{H}(\boldsymbol{x}-\boldsymbol{x}_0)+O(\|\boldsymbol{x}-\boldsymbol{x}_0\|^2) \end{aligned} \tag{2}$$

最后把 LGC 中的流形排序矩阵近似为

$$\boldsymbol{R}\approx\boldsymbol{K}^{-1}+\boldsymbol{G}(\alpha\boldsymbol{I}-\alpha^2\boldsymbol{M}^{\mathrm{T}}\boldsymbol{G})^{-1}\boldsymbol{G}^{\mathrm{T}} \tag{3}$$

式中：$\boldsymbol{K}$ 为对角矩阵；$\boldsymbol{M}$ 和 $\boldsymbol{G}$ 均为 $n\times(d+1)$ 矩阵；$d$ 为样本维度（通常远小于样本数量 $n$ ）。此时可以采用 Woodbury 求矩阵逆公式（即 $(\boldsymbol{A}\boldsymbol{A}^{\mathrm{T}}+\boldsymbol{D})^{-1}=\boldsymbol{D}^{-1}-\boldsymbol{D}^{-1}\boldsymbol{A}(\boldsymbol{I}+\boldsymbol{A}^{\mathrm{T}}\boldsymbol{D}^{-1}\boldsymbol{A})^{-1}\boldsymbol{A}^{\mathrm{T}}\boldsymbol{D}^{-1}$ ）把 $n\times n$ 的矩阵求逆转化为更小的 $(d+1)\times(d+1)$ 矩阵求逆，大大降低了运算量和内存消耗。

## 2.2 后验分布学习算法

后验分布学习算法 (Posterior Distribution Learning, PDL)[91]主要目标是把半监督学习的优势嵌入到传统监督学习中，即充分利用极少量标记样本和大量无标记样本的分布信息在数据输入空间中构建一个稳定可靠的多分类监督学习模型。具体而言，PDL 包括两个组成部分：第一个部分是首先在学习数据集 $X_L$（包括标记样本和部分未标记样本，下标 $L$ 表示学习集）上利用改进的 LGC 算法进行后验估计，第二部分采用加权的多类最小二乘支持向量机对估计出的后验概率进行拟合，如图 1 所示。图中 Posterior Estimator 为后验概率估计器，Density Regressor 为概率密度拟合器，$P(\omega_i\mid X_L)$ 表示给定学习集 $X_L$ 时第 $i$ 类的后延概率分布 $p(\omega_i\mid\boldsymbol{x})$ 表示样本 $\boldsymbol{x}$ 的后延概率。这样做的出发点是通过第一部分的后验估计，算法只利用很少的标记样本和大量未标记样本就能够获得 $X_L$ 中每个数据点的后验概率，而这些数据点的后验概率可以看成是原始数据空间中真实后验概率分布的一组采样；第二部分中使用监督学习算法可以在这些采样点集的基础上对整个数据空间中的后验概率分布进行拟合，从而获得一个较可靠的监督模型来划分数据空间。如果省去第一部分而对数据集进行直接的监督学习，容易出现因为标记样本不足而造成的模型退化，很难获得好的效果，因此第一部分中的后验估计非常重要。

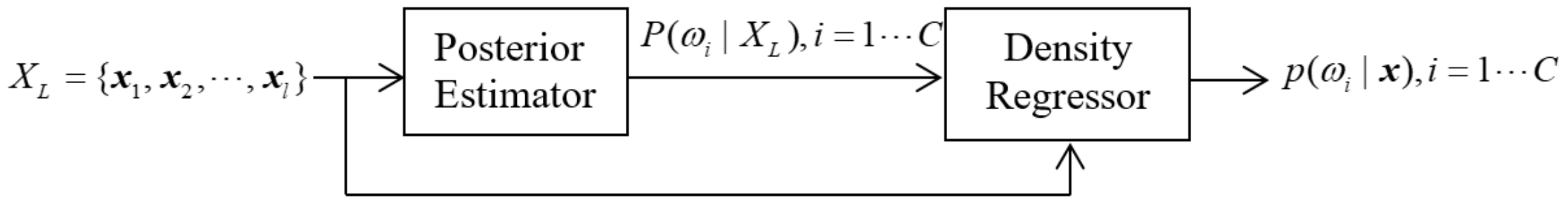

图 1 后验分布学习算法

Fig. 1 Posterior distribution learning algorithm

为了准确地估计样本的后验概率，PLD 采用改进的有约束 LGC 算法在学习数据集上进行标记传播，即利用标记样本同类必连(Must-link)和异类不连(Cannot-link)特性构建有约束的图，然后采用局部传播矩阵取代 LGC 中的全局传播系数，在约束图上学习出每个样本的软标记(Soft Label)并进行归一化后作为学习集中样本的后验概率。概率拟合部分把最小二乘支持向量机[92]由标量函数拟合拓展为向量函数拟合，并引入了误差加权来增加算法稳定性，即求解如下优化问题进行后验概率拟合：

$$\min J(\boldsymbol{A},\boldsymbol{b},\boldsymbol{E}) = \frac{1}{2}\|\boldsymbol{A}\|^2 + \frac{1}{2}\gamma\sum_{j=1}^{n} v_j \|\boldsymbol{e}_j\|^2$$
$$\text{s.t.} \quad F(\boldsymbol{x}_j) = \boldsymbol{A}^{\mathrm{T}}\varphi(\boldsymbol{x}_j) + \boldsymbol{b} + \boldsymbol{e}_j, \quad j = 1,2,\cdots,n \tag{4}$$

式中：$\boldsymbol{A},\boldsymbol{b}$ 分别为系数矩阵和偏移向量；$\boldsymbol{E} = [\boldsymbol{e}_1 \quad \boldsymbol{e}_2 \quad \cdots \quad \boldsymbol{e}_n]$ 为每个样本对应的拟合误差；$F(x)$ 为样本 $x$ 对应的后验概率向量；$\varphi(\bullet)$ 为未知特征映射函数，$\gamma$ 为正则化系数，$v_j$ 为第 $j$ 项所对应的误差权重。注意这里因为映射函数可以是任意维度的非线性函数，因此式(4)不同于传统的岭回归(ridge regression)。式(4)最终可推导为一个线性方程组来求解，因此运算效率较高，而且具有全局最优解。求解后的 $F(\boldsymbol{x})$ 即为所需的监督分类函数，对应给定的任何样本可以直接计算出其对应于每个类的后验概率。

## 2.3　流形 $k$ 近邻半监督学习算法

传统的 $k$ 近邻学习算法具有许多优点，如算法直观易懂、便于实现，参数少易于使用，有理论分析保证，性能可靠等，因此在许多领域有着广泛的应用。但 $k$ 近邻算法对非线性流形分布的数据很难取得好的效果，尤其是在有标记样本（训练样本）很少的情况下，分类精度往往非常差。

流形 $k$ 近邻半监督学习[93]首先采用标记样本约束的无限疲劳随机游走模型计算出具有全局特性的流形相似性度量，然后采用加权的 $k$ 近邻来对无标记样本进行分类。同时，为了解决半监督学习在线分类(即测试样本不在已有的数据集中,而是挨个出现)效率低的问题，流形 $k$ 近邻使用两步优化的局部权值重构算法来计算新样本的流形相似性度量。具体而言，在构建图的过程中，如果两个标记样本来自于同一类，那么它们间的相似性就设置为 1；反之，则设置为 0。然后根据下式计算出无限疲劳随机游走矩阵：

$$\bar{\boldsymbol{P}} = \sum_{k=0}^{\infty}(\alpha\boldsymbol{P})^k = (\boldsymbol{I} - \alpha\boldsymbol{P})^{-1} \tag{5}$$

式中：$0 < \alpha < 1$；$\boldsymbol{P} = \boldsymbol{D}^{-1}\boldsymbol{W}$ 为图上的单步随机游走矩阵，$\boldsymbol{W}$ 为图的邻接矩阵，$\boldsymbol{D}$ 为对角矩阵，其对角元为 $\boldsymbol{W}$ 对应行的和。矩阵 $\bar{\boldsymbol{P}}$ 的第 $(i,j)$ 个元素就是样本 $\boldsymbol{x}_i$ 和 $\boldsymbol{x}_j$ 之间的流形相似性度量，以此为权值就可以使用加权 $k$ 近邻算法对样本进行分类。

对于在线分类情况，新来的测试样本 $\boldsymbol{x}$ 不包括在已有的未标记样本之中，因此如果采用式(5)计算其流形相似性度量，则单个样本的计算开销会很大。为了解决这个问题，流形 $k$ 近邻半监督学习首先从已有的样本中找到测试样本的 $k$ 个近邻放在矩阵 $\boldsymbol{X}_k$ 中，然后通过如下优化问题求得邻域重建权值 $\boldsymbol{z}$：

$$\min_{\boldsymbol{z}\in\mathrm{R}^k} \|\boldsymbol{x} - \boldsymbol{X}_k\boldsymbol{z}\|^2$$
$$\text{s.t.} \quad \boldsymbol{z} \geq \boldsymbol{0}, \ \boldsymbol{z}^{\mathrm{T}}\boldsymbol{e} = 1 \tag{6}$$

式中：$\boldsymbol{e}$ 为全 1 向量。之后把 $\boldsymbol{x}$ 的 $k$ 个近邻的流形度量放在矩阵 $\boldsymbol{W}_k$ 中，通过如下优化问题重构出 $\boldsymbol{x}$ 的流形相似性度量 $\boldsymbol{w}$：

$$\min_{\boldsymbol{w}\in\mathrm{R}^n} \|\boldsymbol{w} - \boldsymbol{W}_k\boldsymbol{z}\|^2$$
$$\text{s.t.} \quad \boldsymbol{w} \geq \boldsymbol{0} \tag{7}$$

这两个优化问题都是二次凸优化问题，利用已有优化算法可以很方便地找到其全局最优解。

## 2.4　变形拉普拉斯矩阵的半监督学习

为了提高算法的抗噪声能力、降低歧义的桥接点给算法性能所带来的不良影响，文献[94]中提出了一种基于变形拉普拉斯矩阵的半监督学习算法。算法首先构造一个 $k$ 近邻图，并计算其拉普拉斯矩阵 $\boldsymbol{L} = \boldsymbol{D} - \boldsymbol{W}$。假设 $f$ 是我们要学习的分类判别软标记向量，对于二分类情况（这里为了方便说明，但实际算法本身无此约束，可多分类），如果用+1、0、−1 分别

表示正样本、无标记样本和负样本，则算法出发点是在目标函数的正则化中除了采用通用的全局流形光滑项 $\boldsymbol{f}^{\mathrm{T}}\boldsymbol{L}\boldsymbol{f}$ 以外，再额外增加一个反映每个样本局部光滑特性的惩罚项：

$$\boldsymbol{f}^{\mathrm{T}}\left(\boldsymbol{I}-\boldsymbol{D}/v\right)\boldsymbol{f}=\sum_{i=1}^{n}\left(1-D_{ii}/v\right)f_i^2 \tag{8}$$

式中：$v$ 是图的体积，即每个节点的度 (Degree) $D_{ii}$ 之和。分析上式右边可以看出，如果某个节点的度较大（对应于类内部的点），那么括号中的系数会较小，从而对标记函数 $f_i$ 施加的惩罚也会较小，则 $f_i^2$ 会有较大的值，即分类置信度较高；类似的分析也可以知道，对于度较小的点（对应于类边缘的点以及歧义的桥接点），$f_i^2$ 会有较小的值，分类的置信度较低。如果假设前 $l$ 个样本为标记样本，则变形拉普拉斯矩阵的半监督学习最终通过求解如下目标函数来实现：

$$\min_{\boldsymbol{f}}\sum_{i=1}^{l}\left(f_i-y_i\right)^2+\beta\boldsymbol{f}^{\mathrm{T}}\boldsymbol{L}\boldsymbol{f}+\gamma\boldsymbol{f}^{\mathrm{T}}\left(\boldsymbol{I}-\boldsymbol{D}/v\right)\boldsymbol{f} \tag{9}$$

式中：$\beta,\gamma$ 为正则化系数。该优化问题为凸优化，且可以求得其最优解的闭合表达式。为了解决算法在线测试的低效率问题，可以把算法拓展到希尔伯特再生核空间中学习出一个归纳模型：

$$\min_{\boldsymbol{f}}\alpha\left\|\boldsymbol{f}\right\|_{\mathcal{H}}^2+\sum_{i=1}^{l}\left(f_i-y_i\right)^2+\beta\boldsymbol{f}^{\mathrm{T}}\boldsymbol{L}\boldsymbol{f}+\gamma\boldsymbol{f}^{\mathrm{T}}\left(\boldsymbol{I}-\boldsymbol{D}/v\right)\boldsymbol{f} \tag{10}$$

式中：第一项对应 $\boldsymbol{f}$ 在希尔伯特再生核空间中的范数平方。利用拓展的表示定理(Generalized Representer Theorem)，上式的全局最优解亦可求得。此外，文献[94]中同时对算法的稳定性和泛化能力给出了严格的证明。

## 2.5　基于菲克定律的半监督学习

物理学中的菲克第一定律描述了物质在具有浓度差的介质中扩散的规律，其通量(Flux) $J$ 与物质量浓度 $\rho$ 之间存在如下关系：

$$J=-\gamma\frac{\mathrm{d}\rho}{\mathrm{d}r} \tag{11}$$

式中：$\gamma$ 为扩散系数；$r$ 为扩散距离。在样本空间中，如果把标记样本看作是信息源的话，那么类别标记信息的含量（即单个样本所携带的类别信息量）也就是物质量浓度 $\rho$，此时分类问题可以看作是类别标记信息量 $\rho$ 从标记样本 $x_i$ 到无标记样本 $x_j$ 的“扩散”过程，因此可以使用菲克第一定律来刻画：

$$J\approx-\gamma\frac{\rho_i-\rho_j}{r_{ij}} \tag{12}$$

式中：$r_{ij}$ 表示样本 $x_i$ 到标记样本 $x_j$ 之间的距离；$\rho_i$ 和 $\rho_j$ 分别为样本 $x_i$ 和样本 $x_j$ 所拥有的类别信息量。为此，文献[95]中提出一种基于菲克定律的半监督学习算法如下：假设 $t$ 时刻的标记信息向量为 $\boldsymbol{f}^{(t)}$（每个元素为对应的样本所含信息量 $\rho$），初始状态向量为 $\boldsymbol{y}$（用+1、0、−1 分别表示正样本、无标记样本和负样本），那么 $t+1$ 时刻的标记信息向量则可使用下式求得：

$$\boldsymbol{f}^{(t+1)}=\alpha\boldsymbol{P}\boldsymbol{f}^{(t)}+\left(1-\alpha\right)\boldsymbol{y} \tag{13}$$

式中：$\boldsymbol{P}$ 是由菲克第一定律导出的扩散矩阵，是一个非负随机矩阵；参数 $0<a<1$ 控制初始状态的权重。通过理论分析可以证明式(13)收敛为

$$\boldsymbol{f}^{*}=\left(1-\alpha\right)\left(\boldsymbol{I}-\alpha\boldsymbol{P}\right)^{-1}\boldsymbol{y} \tag{14}$$

因此信息传递过程可以使用简洁的表达式一步求解，操作方便。同时，对于多分类情况，注意到式(14)对于一组确定的样本，扩散矩阵不变，因此只需把初始状态向量 $y$ 换成多类的初

始状态矩阵即可。相比较已有算法（如 LGC 算法），该算法具有收敛速度快和参数不敏感等优势。

## 2.6 基于导学与导教的半监督学习

以上介绍的算法中，所有样本在学习过程中都是同等对待，这样做的缺点是算法缺乏处理有歧义但关键的数据点（比如桥接点）的能力，容易导致不准确的传播。基于导学与导教的半监督学习[96]通过考查未标记样本的可靠性和可区分性而将未标记样本赋予不同的分类难度，采用导学（即教师）和导教（即学生）两个算法交替学习来优化传播过程，从而实现未标记样本按照由易到难的顺序逐步被分类，以提高算法的分类精度。

具体而言，假设标记样本集合和无标记样本集合分别为 $L$ 和 $U$，对应的标记向量分别为 $\boldsymbol{y}_L$ 和 $\boldsymbol{y}_U$。教师算法首先选择一组可靠性高、区分度大的样本集 $S$，然后学生算法对 $S$ 中的样本进行分类并把结果反馈给教师算法，以决定下次教师算法选取未标记样本的数量。教师算法和学生算法按照如此过程交替进行直到所有未标记样本被分类完成。假设第 $t$ 次教师算法选择的未标记样本集为 $S^{(t)}$，则经过学生算法学习后把 $S^{(t)}$ 中的样本进行分类，并归入到标记样本集中，即 $S^{(t)}$, $U^{(t+1)}=U^{(t)}-S^{(t)} \cup L^{(t+1)}=L^{(t)}$，而初始时 $L^{(0)}=L,\ U^{(0)}=U$。具体算法设计时，教师算法按照如下原则为学生算法选择一组相对容易的样本 $S^*$ 来学习：

$$S^* = \arg\max_{S\subset U} R(S)+D(S) \quad \text{s.t.} \quad |S|=s \tag{15}$$

式中：$R(S)$ 和 $D(S)$ 分别表示 $S$ 中样本的可靠度和区分度；$|S|=s$ 表示集合 $S$ 的大小为 $s$，其数值根据学生学习选结果的评估函数确定，稍后给出。求解时，$R(S)$ 采用条件熵定义为 $R(S)=-H(\boldsymbol{y}_S \mid \boldsymbol{y}_L)=H(\boldsymbol{y}_L)-H(\boldsymbol{y}_{S\cup L})$，而 $D(S)$ 则定义为 $S$ 中的样本 $\boldsymbol{x}$ 与最近的两个类平均通勤时间差 $D(S)=\sum_{\boldsymbol{x}\in S} T(\boldsymbol{x},C_1)-T(\boldsymbol{x},C_2)$，其中 $C_1$ 和 $C_2$ 表示距离 $S$ 中的样本 $\boldsymbol{x}$ 最近的两个类样本集合，而样本 $\boldsymbol{x}$ 到集合 $B$ 的通勤时间 $T(\boldsymbol{x},B)$ 定义为 $\boldsymbol{x}$ 和集合 $B$ 的所有样本间通勤时间均值。两个样本 $\boldsymbol{x}_i$ 和 $\boldsymbol{x}_j$ 的通勤时间是随机游走从 $\boldsymbol{x}_i$ 到达 $\boldsymbol{x}_j$ 然后再返回 $\boldsymbol{x}_i$ 的期望步数。

学生算法则要对给定的未标记样本集合 $S$ 中的样本进行分类，同时把结果反馈给教师算法以便于进行下一个回合的样本筛选。假设第 $t$ 次选择的样本集合为 $S^{(t)}$，学生算法采用如下模型进行分类：

$$\boldsymbol{F}_i^{(t)} = \begin{cases} \boldsymbol{F}_i^{(0)}, & \boldsymbol{x}_i \in L^{(0)} \\ \boldsymbol{P}_{i\cdot}\boldsymbol{F}^{(t-1)}, & \boldsymbol{x}_i \in S^{(1:t-1)} \bigcup S^{(t)} \end{cases} \tag{16}$$

式中：$\boldsymbol{P}_{i\cdot}$ 表示随机游走矩阵 $\boldsymbol{P}=\boldsymbol{D}^{-1}\boldsymbol{W}$ 的第 $i$ 行，$S^{(1:t-1)}$ 表示第 $1,2,\cdots,t-1$ 次选取的样本集合，$L^{(0)}$ 为初始的标记样本集合。对于学习效果的评估，可定义学习函数来量化衡量，例如：

$$g(\boldsymbol{F}_{S^{(t)}}) = \frac{2}{1+\exp\left[-\gamma\left(\left\|\boldsymbol{F}_{S^{(t)}}\right\|_F^2 - s^{(t)}/c\right)\right]} - 1 \tag{17}$$

式中：$\boldsymbol{F}_{S^{(t)}}$ 表示学生算法在第 $t$ 次选择的样本集合 $S^{(t)}$ 上学习后的分类函数（或标记函数）。基于这个学习函数，下一轮迭代中教师算法所选取的样本数 $s^{(t+1)}$ 可通过公式 $s^{(t+1)}=\left\lceil b^{(t+1)} g(\boldsymbol{F}_{S^{(t)}})\right\rceil$ 计算获得，其中 $b^{(t+1)}$ 是所有与 $L^{(t)}$ 中样本相邻的未标记样本数量，$\lceil\ \rceil$ 表示向上取整。

## 3　半监督学习的应用

半监督学习在许多领域有着重要的应用，为了更好地展示其在解决实际问题中的作用，这里介绍两个典型的应用场景：第一个是遥感图像分类，第二个是图像显著性检测。

遥感图像分类是遥感信息处理的一个重要部分，其主要任务是根据遥感区域的不同地物覆盖类型对多光谱卫星图像中的像素进行分类，从而可以通过卫星图像研究地物覆盖类型的组成和变迁，对生态监测、城市规划、土地使用调研等有重要应用[90]。图 2 给出了一个例子。图 2(a)中显示了新疆昌吉地区的一个小区域遥感图像，图 2(b)是对应地区的多光谱伪彩色图，从中可以看出该区域包括四种典型的地物覆盖类型：两种植被覆盖类型，裸土类型和水类型（包括水库和河流）。

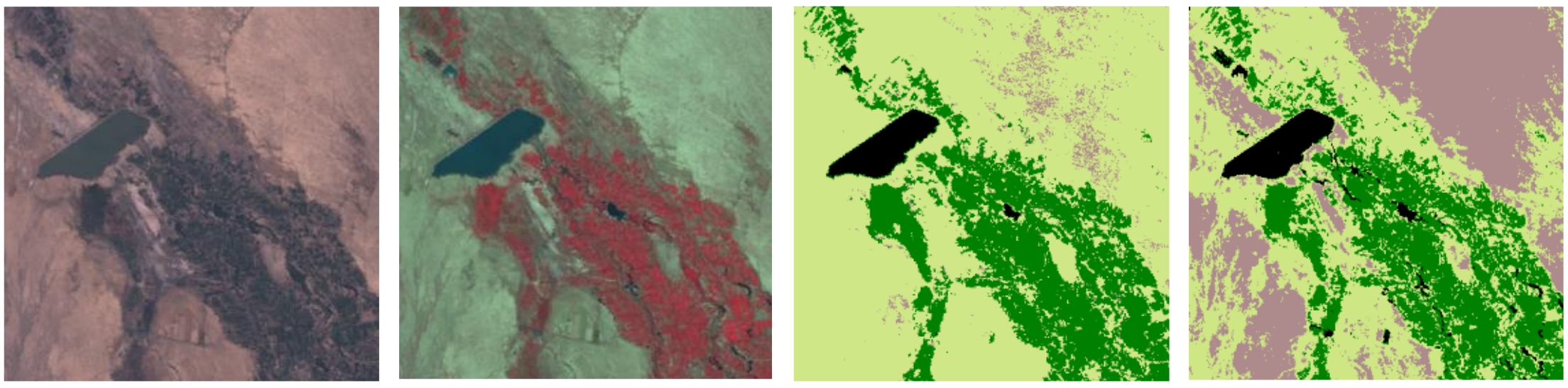

(a) RGB 图像　　(b)伪彩色图像　　(c) 最大似然分类　　(d) 半监督分类

图 2　遥感图像分类应用

Fig. 2　Applications of remote sensing image classification

在实际应用中，因为获取遥感标记数据需要对遥感地区进行实地考察，同时需要专业人员对遥感图像进行解译，因此标记样本的获取成本较高。另外，因为每个类的标记样本很少，所以这些标记样本也无法完全体现出数据的类内多样化和变化范围。这些都给传统的监督学习带来了很大的障碍。图 2(c)显示了最大似然对地物的分类结果，作为对比图 2(d)则是半监督学习遥感图像分类的结果。从中可以看出，在标记样本很有限的情况下，半监督学习分类的结果精度要远好于传统的监督学习。

显著性检测是计算机视觉中的一个重要研究内容，其主要思想里模仿人类的视觉注意力机制，快速地从图像背景中区分出感兴趣的前景目标（而不是简单地对整个图像进行分割）。这在计算机视觉领域有着诸多应用，如目标的自动定位和追踪、目标敏感的图像检索以及内容优先的图像自适应缩放等。如图 3 所示的例子中，需要把图像中的感兴趣目标（这里分别是鸟和狮子）从背景中高亮地标记出来。

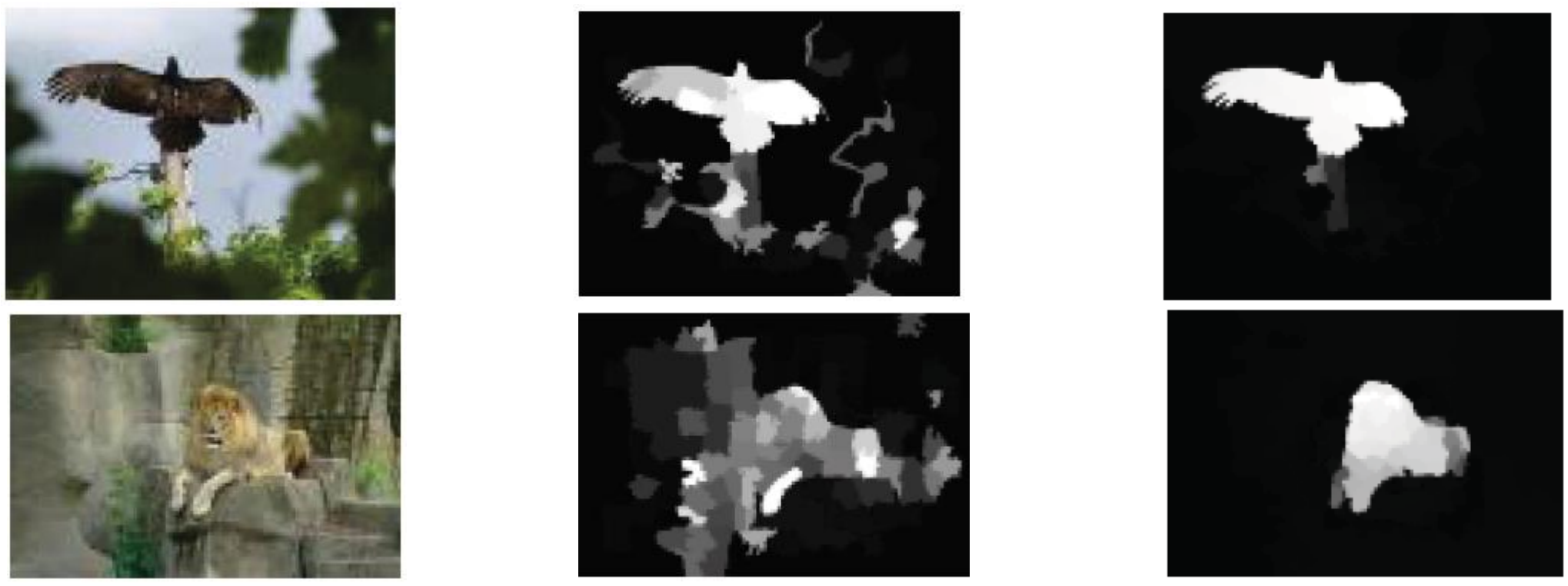

（a）原始图像　　（b）传统算法结果　　（c）半监督学习算法结果

图 3　显著性检测

Fig. 3　Saliency detection results

由于显著性检测过程中极少或者没有手工精确标注的数据，算法只能利用一些常规的先验知识（如图像的边缘像素通常是背景的部分）来对部分像素进行初步标定，因此标记样本很少且质量不高。图 3（b）和（c）分别显示了传统算法和半监督算法的结果。由此可以看出，传统算法很难把复杂背景下的前景目标分离出来，而采用导学与导教的半监督学习算法，通过把图像分割为不同的同质小区域，以及按照由易到难逐步标记的方式能够较好地分离出感兴趣的前景目标。

# 4 结论与讨论

本文总结了半监督学习发展的三个阶段特性并对每个阶段的研究工作做了深入而全面的调研，同时简要介绍了每个阶段的代表性算法和理论，随后介绍了笔者所在实验室在半监督学习研究的最新成果和相关应用。本文对半监督学习的发展过程、主要理论和当前研究重点进行了详细的阐述，为以后的研究和应用提供有益参考。最后需要说明的是半监督学习并不是对所有问题都能取得好的效果，这其中数据分布是否符合半监督学习的模型假设非常重要。模型假设通常有光滑性假设和流形或聚类假设[2, 97]这两个假设含义分别为：若两个样本 $\boldsymbol{x}_1$ 和 $\boldsymbol{x}_2$ 距离很近，则它们更可能来自同一类，即类别标记分布沿着数据分布呈现连续变化；不同类的样本来自于不同的流形上或者类别分布，即不同类别流形或分布间没有（或极少）交叠。在多数的实际应用中，这两个假设通常都能（近似）满足。关于半监督学习的模型假设和更深入、严格的理论分析可以参考文献[98-100]，在此不再详述。

参考文献